# Predicting Human Activities Using Stochastic Grammar

Siyuan Qi[1], Siyuan Huang[1], Ping Wei[2,1], and Song-Chun Zhu[1]
[1]University of California, Los Angeles, USA
[2]Xi'an Jiaotong University, Xi'an, China
{syqi, huangsiyuan}@ucla.edu pingwei@xjtu.edu.cn sczhu@stat.ucla.edu

## Abstract

*This paper presents a novel method to predict future human activities from partially observed RGB-D videos. Human activity prediction is generally difficult due to its non-Markovian property and the rich context between human and environments. We use a stochastic grammar model to capture the compositional structure of events, integrating human actions, objects, and their affordances. We represent the event by a spatial-temporal And-Or graph (ST-AOG). The ST-AOG is composed of a temporal stochastic grammar defined on sub-activities, and spatial graphs representing sub-activities that consist of human actions, objects, and their affordances. Future sub-activities are predicted using the temporal grammar and Earley parsing algorithm. The corresponding action, object, and affordance labels are then inferred accordingly. Extensive experiments are conducted to show the effectiveness of our model on both semantic event parsing and future activity prediction.*

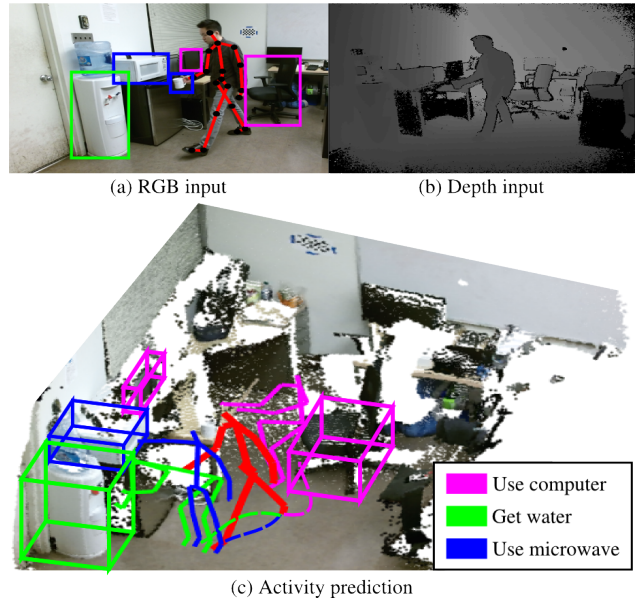

Figure 1: What is he going to do? (a)(b) Input RGB-D video frames. (c) Activity prediction: human action with interacting objects, and object affordances (how the agent will perform the task). The red skeleton is the current observation. The magenta, green and blue skeletons and interacting objects are possible future states.

## 1. Introduction

Consider the image from a video shown in Figure 1(a). A modern computer vision algorithm might reliably detect a human pose and some key objects in the scene: a chair, a monitor, a cup, a microwave and a water cooler. However, we as observers are able to reason beyond the current situation. We can predict what the possible future states are to some extent, and we can even evaluate how strong that belief is – a human can easily predict which state is the most likely future state from Figure 1(c).

The underlying reasoning of the future is more complicated than appearance analysis. The observer needs to understand (i) what happened and what is happening, (ii) what the goal of the agent is, (iii) which object(s) the agent needs to achieve the goal, and (iv) how the agent will perform the task. Based on this rationality, we address the problem of event understanding and human activity prediction from the following two perspectives: (i) a learning algorithm should discover the hierarchical/compositional structure of events, and (ii) an inference algorithm should recover the hierarchical structure given the past observations, and be able to predict the future based on the understanding.

We believe the task of human activity prediction is important for two main reasons. First, the ability to make predictions is key for intelligent systems and robots to perform assistive activities. Second, predicting the future human activities requires deep understanding of human activities. Activity prediction enables the robot to do better task planning. There are attempts that have been made to address this task in both the computer vision [9, 32, 1, 8, 20, 16, 23] and the robotics community [11, 7, 39, 12, 33].

In this paper, we aim to design a model that can (i) learn the hierarchical structure of human activities from videos,

(ii) online infer the current state of the agent and objects while watching a video, and (iii) predict the next states of the agent and objects. Specifically, the state is defined by the action of the agent, the objects that he/she is interacting with and their *affordances* [5], *i.e.* how the objects are being used.

The challenge is three-fold: (i) we need to model the hierarchical structure where the Markov property does not hold. Consider two scenarios: an agent is cleaning the microwave or microwaving food. Whether or not the agent will open the microwave again does not depend on the fact that the agent closed the microwave, but depends on whether or not there is food inside. (ii) Human activities are jointly defined by the human action, the interacting objects, and their affordances. The model needs to capture the spatial-temporal context for event parsing. (iii) We need to predict the human activity from a large future state space.

Inspired by computational linguistics and some recent work in computer vision, we propose a graphical model to represent human activities in a *spatial-temporal And-Or graph* (ST-AOG), which is composed of a spatial And-Or graph (S-AOG) and a temporal And-Or graph (T-AOG). The T-AOG is a stochastic grammar, whose terminal nodes are the root nodes of the spatial graph representing sub-activities. It models the hierarchical structure of human activities and takes the advantage of existing computational linguistic algorithms for symbolic prediction. The S-AOG has child nodes representing a human action, objects, and object affordances. The S-AOG together with T-AOG captures the rich context. For future activity prediction, we first symbolically predict the next sub-activity using the T-AOG, and then predict the human actions and object affordances based on current parsing and sampled future states.

### 1.1. Related Work

**Activity recognition** receives significant attention in recent years, and efforts have been made to detect long-term, complicated activities from videos. A number of methods have been proposed to model the high-level temporal structure of low-level features extracted from video [27, 13, 18, 4, 15]. Some other approaches represent complex activities as collections of attributes [17, 24, 22, 3]. As a recent progress, another stream of work incorporates *object affordances* into activity recognition: Koppula, Gupta and Saxena [10] proposed a model incorporating object affordances that detects and predicts human activities; Wei *et al.* [34] proposed a 4D human-object interaction model for event recognition. We seek to extend this to predict the future activities.

**Future activity prediction** is a relatively new domain in computer vision. [37, 23, 9, 7, 1, 39, 12, 33, 20, 29, 16, 35] predict human trajectories/actions in various settings including complex indoor/outdoor scenes and crowded spaces. Walker *et al.* [32] predicted not only the future motions in the scene but also the visual appearances. In some recent work, Koppula *et al.* [11] used an anticipatory temporal conditional random field to model the spatial-temporal relations through object affordances. Jain *et al.* [8] proposed structural-RNN as a generic method to combine high-level spatial-temporal graphs and recurrent neural networks, which is a typical example that takes advantage of both graphical models and deep learning.

**Grammar models** have been adopted in computer vision and robotics for various tasks. Pei *et al.* [20] unsupervisedly learned a temporal grammar for video parsing. Holtzen *et al.* [7] addressed human intent inference by employing a hierarchical task model. Xiong *et al.* [36] incorporated a spatial, temporal and causal model for robot learning. Gupta *et al.* [6] learned a visually grounded storyline from videos. Grammar-based methods show effectiveness on tasks that have inherent compositional structures.

### 1.2. Contributions

In comparison with the above methods, we make the following contributions:

- We propose a spatial-temporal And-Or graph for human activity understanding to incorporate the hierarchical temporal structure and the rich context captured by actions, objects, and affordances.

- We propose an algorithm for jointly segmenting and parsing the past observations in an online fashion by dynamic programming.

- We propose a novel algorithm to predict the future human activities. Extensive experiments are conducted to show the effectiveness of our approach by evaluating the classification accuracy of actions and affordances.

## 2. Representation

In this section, we introduce the model we propose to represent an activity and define the variables that will be used in the following sections.

An And-Or graph (AOG) is a stochastic context free grammar (SCSG) which represents the hierarchical decompositions from events (top level) to human actions, affordances and objects (bottom level) by a set of terminal and non-terminal nodes. The terminal nodes represent our observation (*e.g.* human and objects in a image). The non-terminal nodes $V_{NT} = V^{And} \cup V^{Or}$ encode the grammar rules. An **And-node** $V^{And}$ represents a decomposition of a large entity (*e.g.* a microwaving-food event) into its constituents (*e.g.* sub-activities such as opening microwave, putting in food). An Or-node $V^{Or}$ represents the possibilities of alternative choices (*e.g.* we can either put a cup or put a sandwich into the microwave). For an And-node $v \in V^{And}$, an **And rule** is defined as a deterministic decomposition $v \to u_1 \cdot u_2 \cdots u_{n(v)}$. For an Or-node $v \in V^{And}$, an **Or rule** is defined as a switch: $v \to u_1 | u_2 | \cdots | u_{n(v)}$, with $p_1 | p_2 | \cdots | p_{n(v)}$. A parse graph $pg$ is an instantiation of the AOG by selecting child nodes for the Or-nodes.

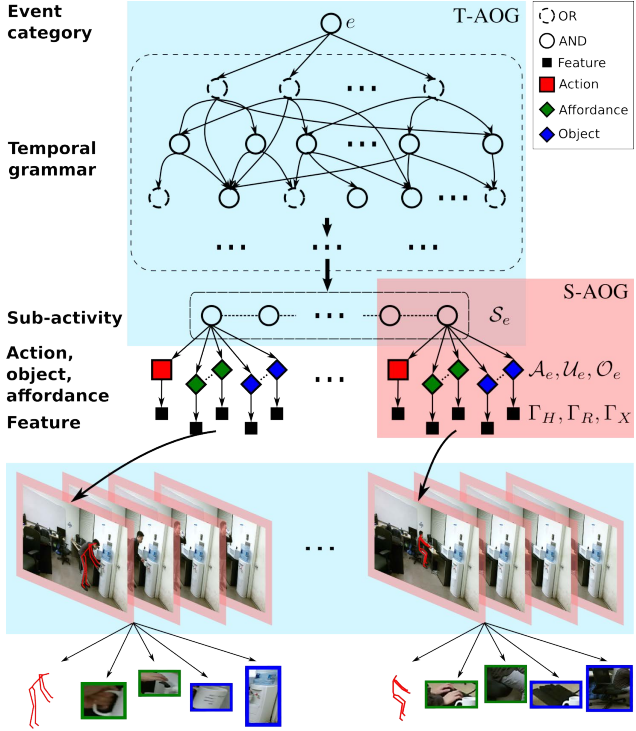

Figure 2: Illustration of the ST-AOG. The sky-blue area indicates the T-AOG, and the coral area indicates the S-AOG. The T-AOG is a temporal grammar in which the root node is the activity and the terminal nodes are sub-activities. The S-AOG represents the state of a scene, including the human action, the interacting objects and their affordances.

Particularly, we represent the task structure as stochastic context free grammar using a spatio-temporal And-Or graph (ST-AOG) as shown in Fig. 2. The ST-AOG can be decomposed into two parts: the spatial AOG (S-AOG) and the temporal AOG (T-AOG). The S-AOG is composed of one And-node expanded into a human action, interacting objects and their affordances, representing the human-object interaction for a video segment. The root And-node of an S-AOG is a sub-activity label. The T-AOG is a temporal grammar, in which the root node is the event and the terminal nodes are sub-activities.

Formally, the ST-AOG of an event $e \in \mathcal{E}$ is denoted by $\mathcal{G}_e = < \mathbb{S}, \mathbb{V}_{NT} \bigcup \mathbb{V}_T, \mathbb{R}, \mathbb{P} >$, where $\mathbb{S}$ is root node. $\mathbb{V}_{NT}$ is the set of non-terminal nodes including the sub-activity labels $\{\mathcal{S}_e\}$. $\mathbb{V}_T = < \{\mathcal{A}_e\}, \{\mathcal{O}_e\}, \{\mathcal{U}_e\} >$ is the set of terminal nodes consist of the human action labels $\{\mathcal{A}_e\}$, the object labels $\{\mathcal{O}_e\}$, and the affordance labels $\{\mathcal{U}_e\}$. $\mathbb{R}$ stands for the production rules, $\mathbb{P}$ represents the probability model defined on the ST-AOG.

For an event in time $[1, T]$, we extract the skeleton features $\Gamma_H$, object features $\Gamma_X$ and the interaction features between the human and the object $\Gamma_R$ from the video $I$. We construct a sequence of parse graphs on $\Gamma = < \Gamma_H, \Gamma_X, \Gamma_R >$, which is defined as $PG = \{pg_t\}_{t=1,\cdots,T}$. $PG$ gives us the label $e$ of the event, and a label sequence $S = \{s_t\}_{t=1,\cdots,T}$ representing the sub-activity labels of all the frames. We obtain the label sequence $H = \{h_t\}$, $O = \{o_t\}$ and $U = \{u_t\}$ for action, affordance and object labels as well. By merging the consecutive frames with the same sub-activity labels, we obtain the temporal parsing of the video, i.e. $\mathcal{T} = \{\gamma_k\}_{k=1,\cdots,K}$ where $\gamma_k = [t_k^1, t_k^2]$ represents a time interval in which the sub-activity remains the same. We use $a^{\gamma_k}$, $o^{\gamma_k}$, and $u^{\gamma_k}$ to denote the action label, object label and affordance label respectively for video segment $I^{\gamma_k}$. Both $a$ and $o$ are vectors, of which lengths are the number of detected objects.

## 3. Probabilistic Formulation

In this section, we introduce the probabilistic model defined on the ST-AOG. Given the extracted action, affordance and object features, the posterior probability of a parse graph sequence $PG$ is defined as:

$$
\begin{aligned}
p(PG|\Gamma, \mathcal{G}_e) &\propto p(\Gamma|PG)p(PG|\mathcal{G}_e) \\
&= p(\Gamma_H, \Gamma_X, \Gamma_R|PG)p(PG|\mathcal{G}_e) \\
&= \underbrace{p(\Gamma_H|PG)}_{\text{action}} \underbrace{p(\Gamma_X|PG)}_{\text{object}} \underbrace{p(\Gamma_R|PG)}_{\text{affordance}} \underbrace{p(PG|\mathcal{G}_e)}_{\text{grammar prior}}
\end{aligned}
\quad (1)
$$

The first three terms are likelihood terms for actions, objects, and affordances given a parse graph $PG$. The last term is a prior probability of the parse graph given the grammar $\mathcal{G}$ of event $e$.

### 3.1. Likelihood of Parse Graphs

#### 3.1.1 Action likelihood

We extract the human skeletons from the Kinect sensor as action features. Assuming that the prior probability for different actions $P(A)$ is uniformly distributed, the prior probability for human skeleton $P(\Gamma_H)$ is normally distributed, the likelihood of action features $\Gamma_H$ given a parse graph $PG$ is defined as:

$$
\begin{aligned}
p(\Gamma_H|PG) = p(\Gamma_H|A) &= \frac{p(A|\Gamma_H)P(\Gamma_H)}{P(A)} \\
&\propto p(A|\Gamma_H)P(\Gamma_H) \\
&= \prod_{k=1}^{K} p(A^{\gamma_k}|\Gamma_H^{\gamma_k})P(\Gamma_H^{\gamma_k})
\end{aligned}
\quad (2)
$$

where $p(A^{\gamma_k}|\Gamma_H^{\gamma_k})$ is the detection probability of an action.

### 3.1.2 Object likelihood

We use the images in the object bounding boxes as object features. The likelihood of object features $\Gamma_X$ given a parse graph $PG$ is given by:

$$p(\Gamma_X|PG) = p(\Gamma_X|O) = \frac{p(O|\Gamma_X)P(\Gamma_X)}{P(O)} \propto p(O|\Gamma_X) = \prod_{k=1}^{K} p(O^{\gamma_k}|\Gamma_X^{\gamma_k}) \quad (3)$$

where we assume that both the prior probability for the image $P(\Gamma_X)$ and $P(O)$ for the object class are uniformly distributed. $p(O^{\gamma_k}|\Gamma_X^{\gamma_k})$ is the detection probability of an object.

### 3.1.3 Affordance likelihood

Given a bounding box of an object in a RGB image, we can extract the point cloud from the corresponding depth image. Based on the detected human skeleton and the object point cloud, we can extract the features for human-object interactions, *i.e.* the distance between the objects and each skeleton joint. The likelihood of human-object interaction features $\Gamma_R$ given a parse graph $PG$ is given by:

$$p(\Gamma_R|PG) = p(\Gamma_R|U) = \frac{p(U|\Gamma_R)P(\Gamma_R)}{P(U)} \quad (4)$$

### 3.2. Grammar Prior of Parse Graphs

After combining the consecutive frames with the same sub-activity labels into segments $\mathcal{T} = \{\gamma_k\}_{k=1,\cdots,K}$, the prior probability of a parse graph $PG$ can be computed by:

$$\begin{aligned}
p(PG|\mathcal{G}_e) &= P(A, O, U|e) \\
&= [\prod_{k=1}^{K} p(a^{\gamma_k}, o^{\gamma_k}, u^{\gamma_k}|s^{\gamma_k}, \gamma_k) p(\gamma_k|s^{\gamma_k})] p(S|e) \\
&= [\prod_{k=1}^{K} p(a^{\gamma_k}|s^{\gamma_k}) p(o^{\gamma_k}|s^{\gamma_k}) p(u^{\gamma_k}|s^{\gamma_k}) p(|\gamma_k| |s^{\gamma_k})] p(S|e)
\end{aligned} \quad (5)$$

where $e$ is the root node of $\mathcal{G}_e$, $p(a^{\gamma_k}|s^{\gamma_k})$, $p(o^{\gamma_k}|s^{\gamma_k})$, and $p(u^{\gamma_k}|s^{\gamma_k})$ are probabilities of observing an action $a$, an object $o$, and an affordance $u$ given the sub-activity $s$ respectively. $p(|\gamma_k| |s^{\gamma_k})$ is the probability of the duration of the segment $|\gamma_k|$ in frames given the sub-activity $s$, modeled by a log-normal distribution. The Viterbi parsing likelihood $p(S|e)$ is the probability of the best parse of the data [28], which is obtained after constructing an AOG based on the temporal parsing results of all videos.

## 4. Learning

The learning of the ST-AOG can be decomposed into two main parts: i) learn the symbolic grammar structure

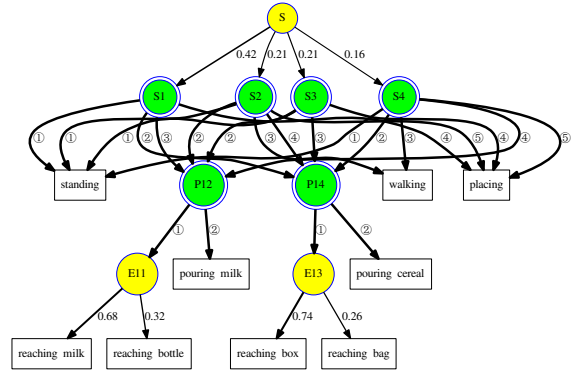

Figure 3: An example of a temporal grammar. The green and yellow nodes are And-nodes and Or-nodes respectively. The numbers on branching edges of Or-nodes represent the branching probability. The circled numbers on edges of And-nodes indicates the temporal order of expansion.

(T-AOG) of each event/task, and ii) learn the parameters $\Theta$ of the ST-AOG, including the branching probabilities of the Or-nodes, the prior distributions of human skeletons and duration of segments.

**Grammar Induction** We used a modified version of the ADIOS (automatic distillation of structure) [26] grammar induction algorithm to learn the event grammar from raw sequential data of symbolic sub-activities and generate the T-AOG whose terminal nodes are sub-activities. The algorithm learns the And-node and Or-nodes by generating significant patterns and equivalent classes. The significant patterns are selected according to a context sensitive criterion defined in terms of local flow quantities in the graph: two probabilities are defined over a search path. One is the right-moving ratio of fan-through (through-going flux of path) to fan-in (incoming flux of paths). The other one, similarly, is the left-going ratio of fan-through to fan-in. The criterion is described in detail in [26].

The algorithm starts by loading the corpus of an activity onto a graph whose vertices are sub-activities, augmented by two special symbols, begin and end. Each event sample is represented by a separate path over the graph. Then it generates candidate patterns by traversing a different search path. In each iteration, it tests the statistical significance of each subpath according to the context sensitive criterion. The significant patterns are recognized as And-nodes. The algorithm then finds the equivalent classes by looking for units that are interchangeable in the given context. The equivalent classes are recognized as Or-nodes. At the end of the iteration, the significant pattern is added to the graph as a new node, replacing the subpaths it subsumes. In our implementation, we favor the shorter significant patterns so that basic grammar units can be captured.

**Parameter learning** The maximum likelihood estimation (MLE) of the branching probabilities of Or-nodes is simply given by the frequency of each alternative choice [38]:

$$\rho_i = \frac{\#(v \to u_i)}{\sum_{j=1}^{n(v)} \#(v \to u_j)} \quad (6)$$

We fit a log-normal distribution for the duration of different sub-activity video segments. A Gaussian distribution is fitted for the human skeletons after aligning the skeletons to a mean pose according to three anchor points, two shoulders and the spine.

## 5. Inference

Given a video as input, our goal is to online predict the human's action, the object he/she is going to interact with, and its affordance, *i.e.* how the object will be used. To accomplish this goal, we first need to parse the past observation, *i.e.* segment the video we have seen and label the human action, objects and affordances for each segment. Then we predict the future states based on our current belief.

### 5.1. Video Parsing

For a single video, we find the parse graph $PG$ for each event $e$ that best explains the extracted features $\Gamma$ by maximizing the posterior probability (1) described in Sec.3:

$$\begin{aligned}PG &= \underset{PG}{\operatorname{argmax}} \, p(PG|\Gamma, \mathcal{G}_e) \\ &= \underset{PG}{\operatorname{argmax}} \, p(\Gamma_H|A)p(\Gamma_X|O)p(\Gamma_R|U)p(A,O,U|e)\end{aligned} \quad (7)$$

Since it is intractable to directly compute the optimal $PG$, we infer the approximately optimal $PG$ by two steps: i) We use a dynamic programming approach to segment the video so that for each segment the action, object and affordance labels remain the same, while maximizing the posterior probability of the labels. ii) After obtaining the video segmentation, we refine the labels according to Eq.7 by Gibbs sampling. Details are described in the following sections.

#### 5.1.1 Segmentation by dynamic programming

To find the video segmentation together with a sequence of labels $S, A, O, U$, we compute the best label $s, a, o, u$ for each video segment with an arbitrary starting frame and end frame and its corresponding probability. Then the segmentation can be obtained by a dynamic programming approach.

For a video segment $I^\gamma$ where $\gamma = [b, f]$ and a given event grammar $\mathcal{G}_e$, we compute the optimal action $a$, object $o$, affordance label $u$, and sub-activity label $s$ by maximizing the posterior probability:

$$\begin{aligned}a, o, u, s &= \underset{a,o,u,s}{\operatorname{argmax}} \, p(a,o,u,s|\Gamma, \gamma) \\ &= \underset{a,o,u,s}{\operatorname{argmax}} \, p(s|a,o,u,\gamma)p(a,o,u|\Gamma^\gamma)\end{aligned} \quad (8)$$

We approximate Equation 8 by first computing $a$, $o$, and $u$:

$$\begin{aligned}a, o, u &= \underset{a,o,u}{\operatorname{argmax}} \, p(a,o,u|\Gamma^\gamma) \\ &= \underset{a,o,u}{\operatorname{argmax}} \, p(a|\Gamma_H^\gamma)p(o|\Gamma_O^\gamma)p(u|\Gamma_X^\gamma)\end{aligned} \quad (9)$$

which is simply the product of detection probabilities of action, objects and affordances. We find out $s$ by:

$$\begin{aligned}s &= \underset{s}{\operatorname{argmax}} \, p(s|a,o,u,\gamma) \\ &\propto \underset{s}{\operatorname{argmax}} \, p(a,o,u,\gamma|s)p(s) \\ &= \underset{s}{\operatorname{argmax}} \, p(a|s)p(o|s)p(u|s)p(|\gamma|\,|s)p(s)\end{aligned} \quad (10)$$

Then the probability of a video until frame $f$ explained by our model is computed by dynamic programming:

$$p(f) = \max_{\substack{b<f \\ a,o,u,s}} p(b)p(a,o,u,s|\Gamma, \gamma=[b,f]) \quad (11)$$

#### 5.1.2 Refine labels by Gibbs Sampling

After obtaining the labels in a bottom-up dynamic programming approach, we refine the labels according to the learned event grammars by Gibbs sampling. For a hypothesized event $e$, we assign the action, affordance and sub-activity labels according to the following probabilities at each iteration:

$$a^{\gamma_k} \sim p(\Gamma_H^{\gamma_k}|a^{\gamma_k})p(a^{\gamma_k}|s^{\gamma_k}) \quad (12)$$

$$u^{\gamma_k} \sim p(\Gamma_R^{\gamma_k}|u^{\gamma_k})p(u^{\gamma_k}|s^{\gamma_k}) \quad (13)$$

$$s^{\gamma_k} \sim p(a^{\gamma_k}, o^{\gamma_k}, u^{\gamma_k}|s^{\gamma_k})p(s^{\gamma_1:\gamma_k}|e) \quad (14)$$

where $s^{\gamma_1:\gamma_k}$ are the labels for the video segments from 1 to $k$. For faster convergence, we use simulated annealing during the Gibbs sampling process.

### 5.2. Human Activity Prediction

Given the current parsing result $PG$ of the observed video sequence, we use the ST-AOG to predict the next sub-activity, action, which object the subject is going to interact with, and how the subject will interact with the object.

Based on the current observation, we predict the future in two stages: i) we symbolically predict the next sub-activities based on the event grammar using an Earley parser [2]. For the current unfinished sub-activity and future sub-activities, we sample the duration in frames based on the learned prior distribution. ii) We predict the human action and affordance labels according to the parse graph and the predicted sub-activity. Assuming that the objects in the scene do not change, we predict the future affordance labels for the existing objects. If we predict that the subject will not interact with an object, the affordance label will be "stationary".

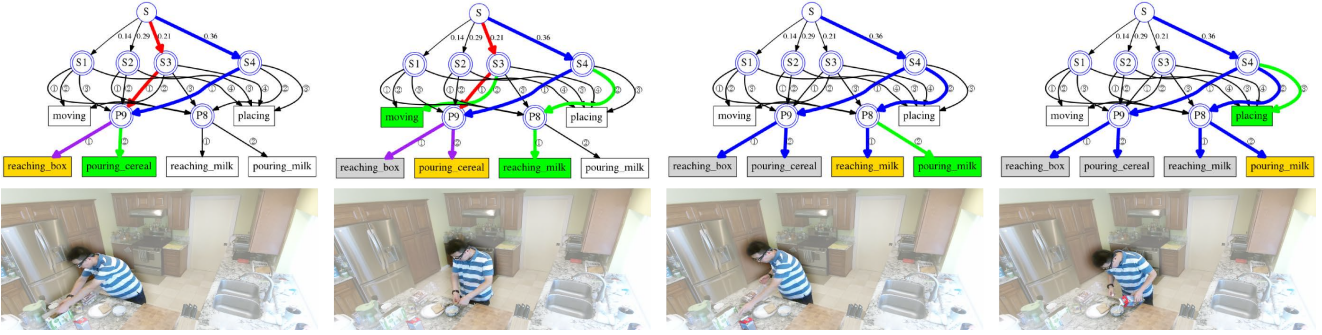

Figure 4: A simplified example illustrating the parsing and symbolic prediction process. In the first two figures, the red edges and blue edges indicates two different parse graphs for the past observations. The purple edges indicate the overlap of the two possible explanations. The red parse graph is eliminated from the third figure. For the terminal nodes, yellow indicates the current observation and green indicates the next possible state(s).

### 5.2.1 Earley parser for sub-activity prediction

We employ an online symbolic prediction algorithm based on the Earley parser to predict the next possible sub-activities in the T-AOG constructed on the sub-activities. Earley parser reads terminal symbols sequentially, creating a set of all pending derivations (states) that are consistent with the input up to the current input terminal symbol. Given the next input symbol, the parser iteratively performs one of three basic operations (prediction, scanning and completion) for each state in the current state set. In our algorithm, we use the current sentence of sub-activities as input into the Earley parser, and scan through all the pending states to find the next possible terminal nodes (sub-activities). Figure 4 shows an illustrative example of the parsing and symbolic prediction process. We then compute the corresponding parsing likelihood for the predicted terminals and sample the sub-activity accordingly.

### 5.2.2 Predict the action and affordance labels

Besides the future sub-activities, we are interested in predicting the future action and affordance labels in a similar manner of event parsing. The difficulty is that we only have the current observation, and we cannot compute the likelihood of the predicted parse graphs of the future. Therefore, to predict the future labels, we propose to sample the future observations (actions and object positions) based on the current observation, and find the best interpretation of the entire observation sequence. Suppose we have the observation of the past $t$ frames and represent the past in a sequence of parse graphs $PG^t$. For a future duration of $d$ frames, we predict the labels by maximizing the posterior probability of $PG^{t+d}$ based on the current observation $\Gamma^t$:

$$\begin{aligned}
p(PG^{t+d}|\Gamma^t) &= \int_{\Gamma^{t:t+d}} p(PG^{t+d}, \Gamma^{t:t+d}|\Gamma^t) \\
&= \int_{\Gamma^{t:t+d}} p(PG^{t+d}|\Gamma^{t:t+d}, \Gamma^t) p(\Gamma^{t:t+d}|\Gamma^t) \\
&= \int_{\Gamma^{t:t+d}} p(PG^{t+d}|\Gamma^{t+d}) p(\Gamma^{t:t+d}|\Gamma^t)
\end{aligned}$$
(15)

The intuition is we compute a joint distribution of the future observation and future parse graphs, and take the marginal distribution as our prediction of the future parse graphs. We use Monte Carlo integration to approximate this probability:

$$p(PG^{t+d}|\Gamma^t) \approx \frac{V}{N} \sum_{i=1}^{N} p(PG^{t+d}|\Gamma_i^{t:t+d}, \Gamma^t) p(\Gamma_i^{t:t+d}|\Gamma^t)$$
(16)

From the current observation, we sample the future human skeleton joint positions and object positions based on the current moving velocities with a Gaussian noise. Then the prediction is obtained by:

$$\begin{aligned}
PG^* &= \underset{PG^{t+d}}{\mathrm{argmax}}\, p(PG^{t+d}|\Gamma^t) \\
&= \underset{PG^{t+d}}{\mathrm{argmax}}\, \frac{V}{N} \sum_{i=1}^{N} p(PG^{t+d}|\Gamma_i^{t:t+d}, \Gamma^t) p(\Gamma_i^{t:t+d}|\Gamma^t) \\
&= \underset{PG^{t+d}}{\mathrm{argmax}} \sum_{i=1}^{N} p(PG^{t+d}|\Gamma_i^{t:t+d}, \Gamma^t) p(\Gamma_i^{t:t+d}|\Gamma^t)
\end{aligned}$$
(17)

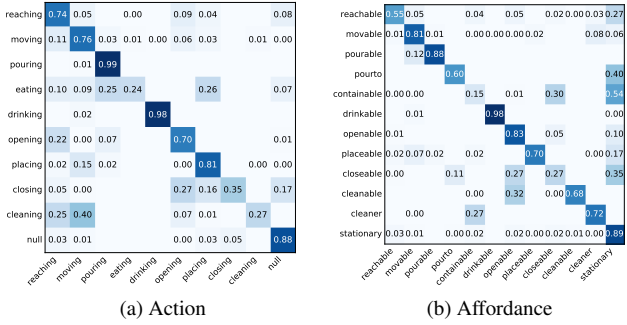

(a) Action  (b) Affordance

Figure 5: Confusion matrices of detection results

|  | Action | | | | Affordance | | | |
|---|---|---|---|---|---|---|---|---|
|  | Micro | Macro | | | Micro | Macro | | |
|  | P/R | Prec. | Recall | F1-score | P/R | Prec. | Recall | F1-score |
| chance | 10.0 | 10.0 | 10.0 | 10.0 | 8.3 | 8.3 | 8.3 | 8.3 |
| SVM | 28.3 | 23.4 | 20.3 | 21.7 | 33.4 | 28.8 | 22.4 | 25.2 |
| LSTM | 36.4 | 30.2 | 27.7 | 28.9 | 42.3 | 36.5 | 33.4 | 34.9 |
| VGG-16 [25] | 46.6 | 54.0 | 31.7 | 40.0 | — | — | — | — |
| KGS [10] | 68.2 | 71.1 | 62.2 | 66.4 | 83.9 | 75.9 | 64.2 | 69.6 |
| ATCRF [11] | 70.3 | 74.8 | 66.2 | 70.2 | **85.4** | **77.0** | **67.4** | **71.9** |
| ours | **76.5** | **77.0** | **75.2** | **76.1** | 82.4 | 72.1 | 66.8 | 69.3 |

Table 1: Detection results on the CAD-120 dataset

|  | Action | | | | Affordance | | | |
|---|---|---|---|---|---|---|---|---|
|  | Micro | Macro | | | Micro | Macro | | |
|  | P/R | Prec. | Recall | F1-score | P/R | Prec. | Recall | F1-score |
| chance | 10.0 | 10.0 | 10.0 | 10.0 | 8.3 | 8.3 | 8.3 | 8.3 |
| LSTM | 24.1 | 22.6 | 19.5 | 19.0 | 31.2 | 28.5 | 23.4 | 25.7 |
| KGS [10] | 28.6 | – | – | 11.1 | 55.9 | – | – | 11.6 |
| ATCRF [11] | 49.6 | – | – | 40.6 | 67.2 | – | – | 41.4 |
| ours | **55.2** | **56.5** | **56.6** | **56.6** | **73.5** | **58.9** | **53.8** | **56.2** |

Table 2: Prediction results on the CAD-120 dataset

## 6. Experiments and Evaluations

In this section we describe the evaluation of our proposed approach on online parsing and prediction. We perform our experiments on CAD-120 dataset [10]. It has 120 RGB-D videos of four different subjects performing 10 activities, each of which is a sequence of sub-activities involving 10 actions (*e.g.* reaching, opening), and 12 object affordance (*e.g.* reachable, openable) in total. We compare our method with recently proposed methods [11, 10] and several other baselines. [1]

### 6.1. Parsing Results

We parsed the videos frame by frame in an online fashion and evaluate the detection results for the current frame. The model is trained on three subjects and tested on a new subject. Results are obtained by four-fold cross validation by averaging across the folds. We trained the action and affordance detectors using a simple two-layer fully connected neural network based on features similar to [11]. We fine-tuned Faster R-CNN [21] for object detection. We compared our detection results with the following methods: 1) Chance. The labels are chosen randomly. 2) SVM: An SVM trained on our features. 3) LSTM: A two-layer LSTM trained on our features. 4) VGG-16 [25]: Using the image as input, we fine-tuned a VGG-16 network on the action labels. Since the object affordances are evaluated on each object instead of an image (an image can have multiple objects thus can have multiple affordance labels), we only evaluate the performance of action detection. 5) KGS [10]: A Markov random field model where the nodes represent objects and sub-activities, and the edges represent the spatial-temporal relationships. 6) ATCRF [11]: An anticipatory temporal conditional random field that models the spatial-temporal relations through object affordances.

---

[1] In this paper, we use the term "sub-activities" in a complete sense that involves actions, objects, and affordances (e.g. reaching a plate). In CAD-120 vocabulary, the "sub-activity" labels are reaching, moving, etc, which we consider being "actions".

Figure 5 shows the confusion matrix for classifying actions and affordances, and we report the overall micro accuracy, macro precision and macro recall of the detected actions and affordances in Table 1. Our approach outperforms the other methods on action detection, and achieves a comparable performance with ATCRF [11] on affordance detection.

In the experiments, we found that the algorithm is generally capable of improving the low-level detections using joint high-level reasoning. For example, one "stacking objects" video has an input action detection accuracy of 50.9% and affordance detection accuracy of 84.5%. After joint reasoning, the output action detection accuracy raised to 86.7% and affordance detection accuracy raised to 87.3%.

### 6.2. Prediction Results

We report the frame-wise accuracy of prediction on actions and affordances over 3 seconds in the future (using frame rate of 14Hz as reported in [11]). Table 2 shows the comparisons between our approach and other methods. We achieved a better performance for all predictions even though the detection result is not the best.

One major difficulty in the prediction process is that the parsed sub-activities are often grammatically incorrect due to the wrong detections. In the cases where the temporal structure (segmentation) is roughly correct, the Gibbs sampling described in Section 5.1.2 can correct the wrong labels. However, there are cases when noisy low-level detection results bring challenge to our predictions. There exist work in computational linguistics [19, 30, 31] that address the problem of parsing grammatically incorrect sentences. In our implementation, we sampled a corpus of different activities and find the nearest sentence to the current observation from the corpus by computing the longest common subsequence between sentences. Predictions are then made based on the parsing result of the nearest sentence.

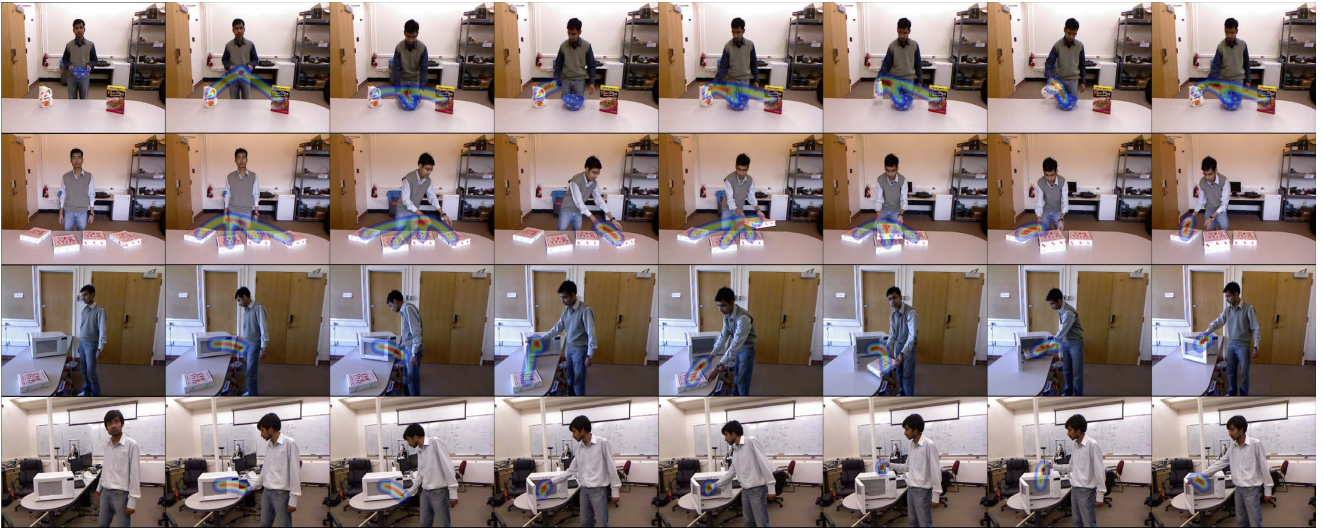

Figure 6: Qualitative results of affordance prediction. Top to bottom: making cereal, stacking objects, taking food, and microwaving food. The first column shows the start frame of the video.

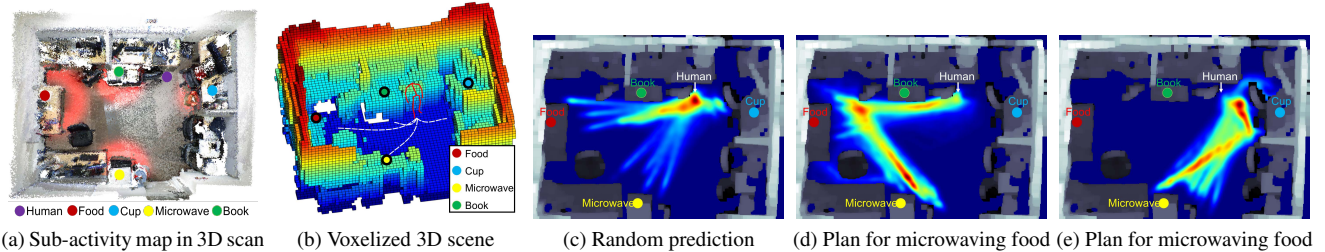

(a) Sub-activity map in 3D scan  (b) Voxelized 3D scene  (c) Random prediction  (d) Plan for microwaving food  (e) Plan for microwaving food

Figure 7: Qualitative results of planning. (a) shows the heat map for human activities. (b) shows the voxelized 3D scene and example trajectories from the human position to sampled target positions. (c)(d)(e) shows the trajectory heat map for random prediction and plans for microwaving food.

### 6.3. Qualitative results

**Prediction** Based on the predicted affordance labels, we can predict which object human is going to interact with. Figure 6 shows the predicted right hand trajectory heat maps within the next one second.

**Task planning** Besides online parsing and prediction of activities, our proposed method can help task planning using the learned T-AOG. Given a 3D scene and a task, we can generate different possible task plans according to the learned grammar. As shown in Figure 7(a), after obtaining a 3D scanned scene, we can compute the human activity heat maps with respect to different labeled objects. The heat maps are computed using the voxelized 3D scene and average poses of actions associated with the objects. Based on the heat map, we can sample target positions for interactions with the objects, and then plan trajectories from the human position to the objects. Figure 7(b) illustrates the possible paths from the human position to the targets in a voxelized scene. Figure 7(c) shows the heat map of trajectories assuming the human randomly select a target object. The trajectories are planned using rapidly-exploring random tree (RRT) [14]. Based on the event grammar, we can also symbolically sample different plans for learned tasks, and plan multiple trajectories. Figure 7(d)(e) show examples of different trajectory heat maps for "microwaving food".

## 7. Conclusion

This paper presents a method for online human activity prediction from RGB-D videos. We modeled the activities using a spatial-temporal And-Or graph (ST-AOG). The results show the effectiveness of our model on both detection and anticipation, as well as how the learned model can be used for robot planning. In the future, we could explore object-part based affordances to learn more fine-grained activities.

## Ackowledgements

This research was supported by grants DARPA XAI project N66001-17-2-4029, ONR MURI project N00014-16-1-2007, and NSF IIS-1423305.

# Supplementary Materials

## 1. Temporal Grammar

### 1.1. Grammar Induction

There are several parameters in our implementation of the ADIOS algorithm:

- $\eta$: threshold of detecting divergence in the ADIOS graph for the right-moving ratio $P_R$ and the left-going ratio $P_L$. In our experiment, this is set to 0.9.

- $\alpha$: significance test threshold for the decrease of $P_R$ and $P_L$.
  our experiment, this is set to 0.1.

- context size: size of the context window used for search for equivalence classes. In our experiment, this is set to 4.

- coverage: minimum overlap for bootstrapping Equivalence classes. Higher values will result in less bootstrapping. In our experiment, this is set to 0.5.

### 1.2. Earley Parser

This section gives an introduction to the Earley parser and how we use the Earley's algorithm to predict the next symbol. In the following descriptions, $\alpha$, $\beta$, and $\gamma$ represent any string of terminals/nonterminals (including the empty string), $X$ and $Y$ represent single nonterminals, and a represents a terminal symbol. We use Earley's dot notation: given a production $X \to \alpha\beta$, the notation $X \to \alpha \cdot \beta$ represents a condition in which $\alpha$ has already been parsed and $\beta$ is expected.

Input position 0 is the position prior to input. Input position n is the position after accepting the nth token. (Informally, input positions can be thought of as locations at token boundaries.)

For every input position, the parser generates a state set. Each state is a tuple $(X \to \alpha \cdot \beta, i)$, consisting of

- The production currently being matched ($X \to \alpha\beta$).

- The current position in that production (represented by the dot)

- The position i in the input at which the matching of this production began: the origin position

The state set at input position $k$ is called $S(k)$. The parser is seeded with $S(0)$ consisting of only the top-level rule. The parser then repeatedly executes three operations: prediction, scanning, and completion.

- Prediction: for every state in $S(k)$ of the form $(X \to \alpha \cdot Y\beta, j)$, where $j$ is the origin position as above, add $(Y \to \cdot\gamma, k)$ to $S(k)$ for every production in the grammar with Y on the left-hand side($Y \to \gamma$).

- Scanning: if a is the next symbol in the input stream, for every state in $S(k)$ of the form $(X \to \alpha \cdot a\beta, j)$, add $(X \to \alpha a \cdot \beta, j)$ to $S(k+1)$.

- Completion: for every state in $S(k)$ of the form $(X \to \gamma\cdot, j)$, find states in $S(j)$ of the form $(Y \to \alpha \cdot X\beta, i)$ and add $(Y \to \alpha X \cdot \beta, i)$ to $S(k)$.

In our prediction for the next possible symbol at current position $t$, we search through the states $S(t)$ of the form $(X \to \alpha \cdot a\beta, j)$, where the first symbol after the current position is a terminal node. The predictions $\Sigma$ are then given by the set of all possible $a$:

$$\Sigma = \{a : \exists s \in S(t), s = (X \to \alpha \cdot a\beta, j)\} \quad (1)$$

The probability of each production is then given by the parsing likelihood of the sentence constructed by appending the predicted state to the current sentence.

### 1.3. Parsing Likelihood

For a grammatically complete sentence $s$, the parsing likelihood is simply the Viterbi likelihood. For a incomplete sentence $e$ of length $k$, the parsing likelihood is given by the sum of all the grammatically possible sentences:

$$p(e) = \sum_{s_{[1:k]}=e} p(s) \quad (2)$$

where $s_{1:k}$ denotes the first $k$ words of a complete sentence $s$, and $p(s)$ is the Viterbi likelihood of $s$.

## 2. Experiments

### 2.1. Feature Design

In our method, we extracted different features for action and affordance detection/prediction.

The action feature is composed of the positions of eleven key joints from the upper body of human skeleton and the relative distances and orientations between each two joints. The affordance feature is concatenated by the action feature, the main position of object point clouds, and the relative distances and orientations between the center of the object and several key skeleton joints such as hand and head.

We extracted both features for each frame of the videos to employ our algorithm.

## 2.2. Compared Methods

We compared with some baseline methods in experiments part.

- Chance. We randomly choose the label for detection and prediction.

- SVM. We treated the task as classification problem and used SVM to detect and predict. We employed multi-class SVM to train detection classifier with action/affordance feature and action/affordance label of current frame. We evaluated the detection performance with the classifier.

- LSTM. We split each video to several segments of length ten. We treated each segment as a sequential input for LSTM with our features and utilized the label of ten frames to train the LSTM. We built a two-layer LSTM with softmax layer on top of it to process the sequential feature and got a sequential output label. We used labels of current ten frames to detect and labels of the next ten corresponding frames in 3 seconds to predict.

- VGG-16. We extracted the human images with the skeleton positions and the camera parameters. Similarly, we fine-tuned VGG-16 network to train the action classifier for detection. Since the object affordances are evaluated on each object instead of an image (an image can have multiple objects thus can have multiple affordance labels), we only evaluate the performance of action detection.

- KGS and ATCRF are introduced in [1] and [2] respectively.